\pdfoutput=1
\RequirePackage{amsmath}         %

\documentclass[runningheads,a4paper]{llncs}

\usepackage{graphicx}            %
\usepackage{amssymb}             %
\usepackage{amsfonts}            %
\usepackage{enumitem}            %
\usepackage{multirow}            %
\usepackage{siunitx}             %
\usepackage{url}                 %
\usepackage{xspace}              %
\usepackage[T1]{fontenc}         %
\usepackage[hidelinks]{hyperref} %

\graphicspath{{images/}}
\DeclareGraphicsExtensions{.pdf,.png,.jpg,.jpeg}

\newcommand{\degreem}{^{\circ}} %

\newcommand{\figlabel}[1]{\label{fig:#1}}
\newcommand{\tablabel}[1]{\label{tab:#1}}

\newcommand{\figref}[1]{Fig.~\ref{fig:#1}\xspace}
\newcommand{\tabref}[1]{Table~\ref{tab:#1}\xspace}

\newcommand{\nop}{NimbRo\protect\nobreakdash-OP\xspace}
\newcommand{\dop}{DARwIn\protect\nobreakdash-OP\xspace}

\newcommand{\nao}{Nao\xspace}
\newcommand{\cm}{CM730\xspace}
\newcommand{\cmnew}{CM740\xspace}
\newcommand{\itwoc}{I\textsuperscript{2}C\xspace}
\newcommand{\igus}{igus\textsuperscript{\tiny\circledR}\xspace}
\newcommand{\iguhop}{igus\textsuperscript{\tiny\circledR}$\!$ Humanoid Open Platform\xspace}
\newcommand{\iguhopp}{igus\textsuperscript{\tiny\circledR} Humanoid Open Platform\xspace}

\newcommand{\term}[1]{\emph{#1}\xspace}
\newcommand{\degree}{$\degreem$\xspace}

\newcommand{\ssection}[1]{\vspace{-1.0ex}\section{#1}\vspace{-1.0ex}}
\newcommand{\ssections}[1]{\vspace{-1.0ex}\section*{#1}\vspace{-1.0ex}}
\newcommand{\ssubsection}[1]{\vspace{-1.0ex}\subsection{#1}\vspace{-1.0ex}}

\usepackage{eso-pic}

\AtBeginDocument{\AddToShipoutPictureFG*{\AtTextUpperLeft{\put(0,\LenToUnit{40pt}){\parbox{\textwidth}{\centering\bfseries
RoboCup 2016: Robot World Cup XX, Lecture Notes in\\Computer Science 9776, Springer, 2017
}}}}}
\begin{document}

\mainmatter

\title{First International HARTING Open Source Prize Winner: The igus Humanoid Open Platform}
\titlerunning{The igus Humanoid Open Platform}

\author{Philipp Allgeuer, Grzegorz Ficht, Hafez Farazi,\\Michael Schreiber and Sven Behnke}
\authorrunning{P. Allgeuer, G. Ficht, H. Farazi, M. Schreiber, and S. Behnke}

\institute{Autonomous Intelligent Systems, Computer Science, Univ.\ of Bonn, Germany\\
\url{{pallgeuer, ficht, hfarazi}@ais.uni-bonn.de}, \url{behnke@cs.uni-bonn.de}\\
\url{http://ais.uni-bonn.de}\vspace{-2ex}}

\maketitle

\begin{abstract}
The use of standard platforms in the field of humanoid robot\-ics can lower the 
entry barrier for new research groups, and accelerate research by the 
facilitation of code sharing. Numerous humanoid standard platforms exist in the 
lower size ranges of up to 60\,cm, but beyond that humanoid robots scale up 
quickly in weight and price, becoming less affordable and more difficult to 
operate, maintain and modify. The \iguhop is an affordable, fully open-source 
platform for humanoid research. At 92\,cm, the robot is capable of acting in an 
environment meant for humans, and is equipped with enough sensors, actuators and 
computing power to support researchers in many fields. The structure of the 
robot is entirely 3D printed, leading to a lightweight and visually appealing 
design. This paper covers the mechanical and electrical aspects of the robot, as 
well as the main features of the corresponding open-source ROS software. At 
RoboCup 2016, the platform was awarded the first International HARTING Open 
Source Prize.
\end{abstract}

\ssection{Introduction}

The field of humanoid robotics is enjoying increasing popularity, with many 
research groups having developed robotic platforms to investigate topics such as 
perception, manipulation and bipedal walking. The entry barrier to such research 
can be significant though, and access to a standard humanoid platform can allow 
for greater focus on research, and facilitates collaboration and code exchange.

The \iguhop, described in this paper, is a collaboration between 
researchers at the University of Bonn and \igus GmbH, a leading manufacturer of 
polymer bearings and energy chains. The platform seeks to close the gap between 
small, albeit affordable, standard humanoid platforms, and larger significantly 
more expensive ones. We designed the platform to be as open, modular, 
maintainable and customisable as possible. The use of almost exclusively 3D 
printed plastic parts for the mechanical components of the robot is a result of 
this mindset, which also simplifies the manufacture of the robots. This allows 
individual parts to be easily modified, reprinted and replaced to extend the 
capabilities of the robot, shown in \figref{P1_teaser}. A demonstration video of 
the \iguhop is available.\footnote{Video: 
\url{https://www.youtube.com/watch?v=RC7ZNXclWWY}}

\begin{figure}[!t]
\parbox{\linewidth}{\centering
\raisebox{8pt}{\includegraphics[width=0.25\linewidth]{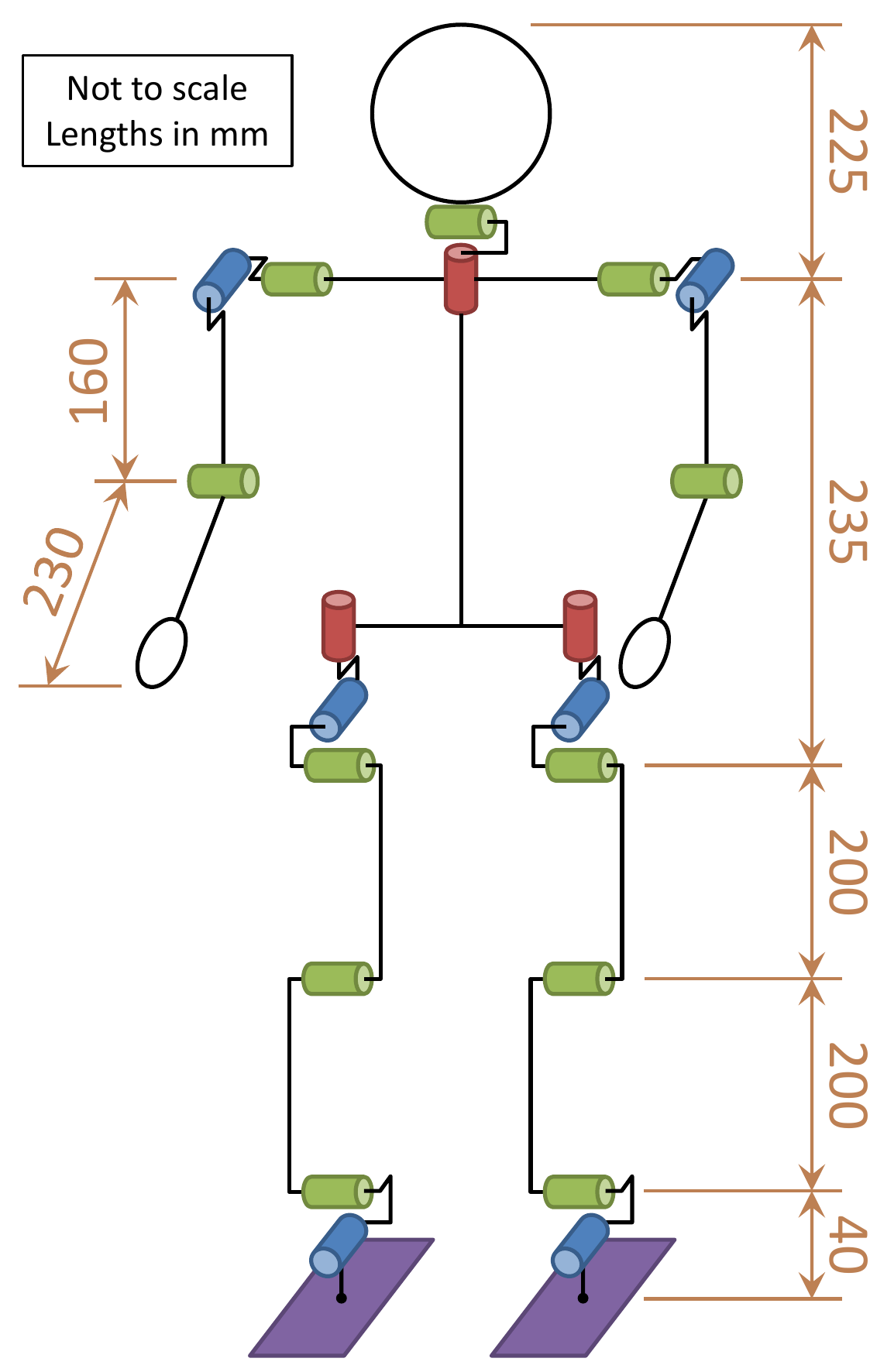}}\hspace{0.04\linewidth}
\includegraphics[height=52mm]{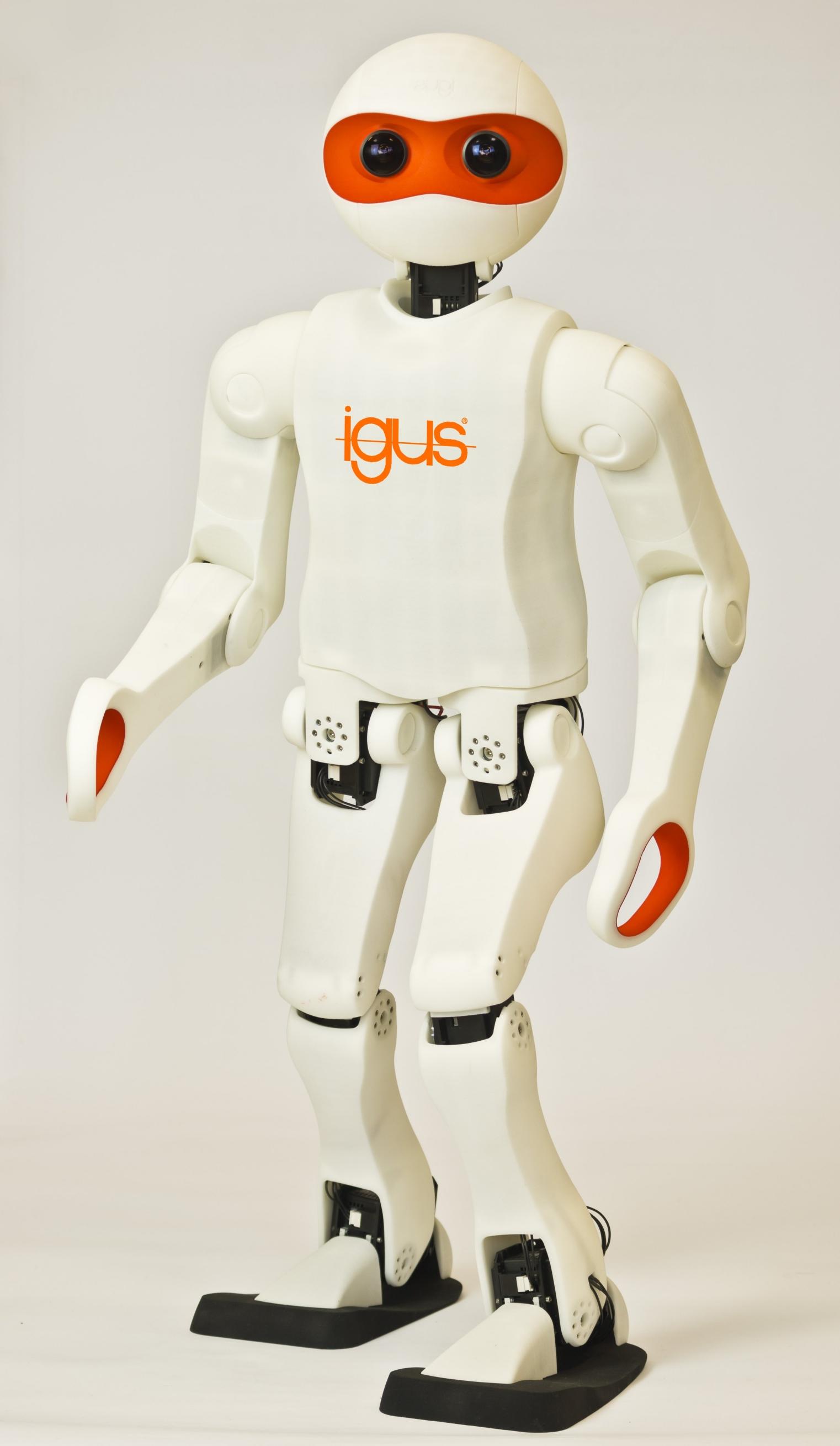}\hspace{0.02\linewidth}
\includegraphics[height=52mm]{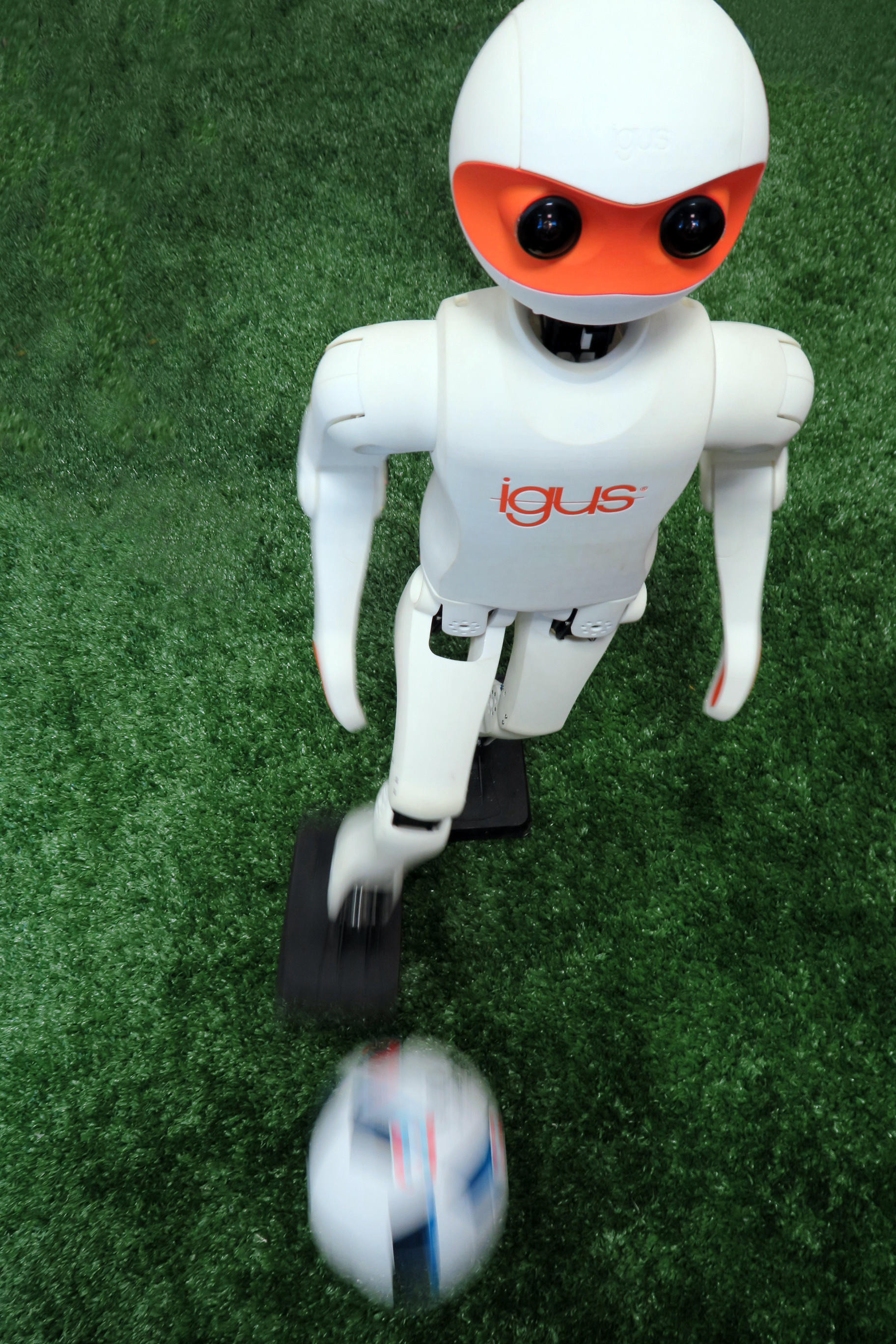}}
\caption{The \iguhop and its kinematic structure.}
\figlabel{P1_teaser}
\end{figure}
\ssection{Related Work}

A number of standard humanoid robot platforms have been developed over the last 
decade, such as for example the \nao robot by Aldebaran Robotics 
\cite{Gouaillier2009}. The \nao comes with a rich set of features, such as a 
variety of available gaits, a programming SDK, and human-machine interaction 
modules. The robot however has a limited scope of use, as it is only \SI{58}{cm} 
tall. Also, as a proprietary product, own hardware repairs and enhancements are 
difficult. Another example is the \dop \cite{Ha2011}, distributed by Robotis. At 
\SI{45.5}{cm}, it is half the size of the \iguhop. The \dop has the benefit of 
being an open platform, but its size remains a limiting factor for its range of 
applications.

Other significantly less widely disseminated robots include the Intel Jimmy 
robot, the Poppy robot from the Inria Flowers Laboratory \cite{Lapeyre2014}, and 
the Jinn-Bot from Jinn-Bot Robotics \& Design GmbH in Switzerland. All of these 
robots are at least in part 3D printed, and the first two are open source. The 
Jimmy robot is intended for social interactions and comes with software based on 
the \dop framework. The Poppy robot is intended for non-autonomous use, and 
features a multi-articulated bio-inspired morphology. Jinn-Bot is built from 
over 90 plastic parts and 24 actuators, making for a complicated build, and is 
controlled by a Java application running on a smartphone mounted in its head. 
Larger humanoid platforms, such as the Asimo \cite{Hirai1998}, HRP 
\cite{Kaneko2009} and Atlas robots, are an order of magnitude more expensive and 
more complicated to operate and maintain. Such large robots are less robust 
because of their complex hardware structure, and require a gantry in normal use. 
These factors limit the use of such robots by most research groups.

\ssection{Hardware Design}

The hardware platform was designed in collaboration with \igus GmbH, which 
engaged a design bureau to create an appealing overall aesthetic appearance. The 
main criteria for the design were the simplicity of manufacture, assembly, 
maintenance and customisation. To satisfy these criteria, a modular design 
approach was used. Due to the 3D printed nature of the robot, parts can be 
modified and replaced with great freedom. A summary of the main hardware 
specifications of the \iguhop is shown in \tabref{P1_specs}. 

\begin{table}[tb]
\renewcommand{\arraystretch}{1.2}
\vspace{-1ex}
\caption{\iguhop specifications}
\vspace{-2ex}
\tablabel{P1_specs}
\centering
\footnotesize
\begin{tabular}{c c c}
\hline
\textbf{Type} & \textbf{Specification} & \textbf{Value}\\
\hline
\multirow{2}{*}{\textbf{General}} & Physical & \SI{92}{cm}, \SI{6.6}{kg}, Polyamide 12 (PA12)\\
& Battery & 4-cell LiPo (\SI{14.8}{V}, \SI{3.8}{Ah}), \SI{15}{}--\SI{30}{\minute}\\
\hline
\multirow{2}{*}{\textbf{PC}} & Product & Gigabyte GB-BXi7-5500, Intel i7-5500U, \SI{2.4}{}--\SI{3.0}{GHz}\\
& Options & \SI{4}{GB} RAM, \SI{120}{GB} SSD, Ethernet, Wi-Fi, Bluetooth\\
\hline
\multirow{2}{*}{\textbf{\cm}} & Microcontroller & STM32F103RE, \SI{512}{KB} Flash, \SI{64}{KB} SRAM\\
& Interfaces & 3$\,\times\,$Buttons, 7$\,\times\,$Status LEDs\\
\hline
\multirow{2}{*}{\textbf{Actuators}} & Total & 8$\,\times\,$MX-64, 12$\,\times\,$MX-106\\
& Per Limb & 2$\,\times\,$MX-64 (head), 3$\,\times\,$MX-64 (arm), 6$\,\times\,$MX-106 (leg)\\
\hline
\multirow{3}{*}{\textbf{Sensors}} & Encoders & \SI{4096}{ticks/rev} per joint axis\\
& IMU & 9-axis (L3G4200D, LIS331DLH, HMC5883L)\\
& Camera & Logitech C905 (720p), with 150\degree FOV wide-angle lens\\
\hline
\end{tabular}
\vspace{-3ex}
\end{table}
\ssubsection{Mechanical Structure}

The plastic shell serves not only for outer appearance, but also as the 
load-bearing frame. This makes the \iguhop very light for its size. Despite its 
low weight, the robot is still very durable and resistant to deformation and 
bending. This is achieved by means of wall thickness modulation in the areas 
more susceptible to damage, in addition to strategic distribution of ribs and 
other strengthening components, printed directly as part of the exoskeleton. 
Utilising the versatile nature of 3D printing, the strengths of the plastic 
parts can be maximised exactly where they are needed, and not unnecessarily so 
in other locations. If a weak spot is identified through practical experience, 
as indeed happened during testing, the parts can be locally strengthened in the 
CAD model without significantly impacting the remaining design.

\ssubsection{Robot Electronics}

The electronics of the platform are built around an Intel i7-5500U processor, 
running a full 64-bit Ubuntu OS. DC power is provided via a power board, where 
external power and a 4-cell Lithium Polymer (LiPo) battery can be connected. The 
PC communicates with a Robotis \cm subcontroller board, whose main purpose is to 
electrically interface the twelve MX-106 and eight MX-64 actuators, all 
connected on a single star topology Dynamixel bus. Due to a number of 
reliability and performance factors, we redesigned and rewrote the firmware of 
the \cm (and \cmnew). This improved bus stability and error tolerance, and 
decreased the time required for the reading out of servo data, while still 
retaining compatibility with the standard Dynamixel protocol. The \cm 
incorporates 3-axis gyroscope and accelerometer sensors, is connected externally 
to an additional 3-axis magnetometer via an \itwoc interface, and also connects 
to an interface board that has three buttons, five LEDs and two RGB LEDs.

Further available external connections to the robot include USB, HDMI, Mini 
DisplayPort, Gigabit Ethernet, IEEE 802.11b/g/n Wi-Fi, and Bluetooth 4.0. The 
\iguhop is nominally equipped with a single 720p Logitech C905 camera behind its 
right eye, fitted with a wide-angle lens. A second camera can be optionally 
mounted behind the left eye for stereo vision.

\ssection{Software}

The ROS middleware was chosen as the basis of the software developed for the 
\iguhop. This fosters modularity, visibility, reusability, and to some degree 
the platform independence. The software was developed with humanoid robot soccer 
in mind, but the platform can be used for virtually any other application. This 
is possible because of the strongly modular way in which the software was 
written, greatly supported by the natural modularity of ROS, and the use of 
plugin schemes.

\ssubsection{Vision}

The camera driver used in the ROS software nominally retrieves images at 
\SI{30}{Hz} in 24bpp BGR format at a resolution of $640\!\times\!480$. For 
further processing, the captured image is converted to the HSV colour space. In 
our target application of soccer, the vision processing tasks include field, 
ball, goal, field line, centre circle and obstacle detection, as illustrated in 
\figref{vision_output} \cite{Farazi2015}. The wide-angle camera used introduces 
significant distortion, which must be compensated when projecting image 
coordinates into egocentric world coordinates. We undistort the image with a 
Newton-Raphson based approach (top right in \figref{vision_output}). This method 
is used to populate a pair of lookup tables that allow constant time distortion 
and undistortion operations at runtime. Further compensation of projection 
errors is performed by calibrating offsets to the position and orientation of 
the camera frame in the head of the robot. This is essential for good projection 
performance (bottom row in \figref{vision_output}), and is done using the 
Nelder-Mead method.

\begin{figure}[!t]
\parbox{\linewidth}{\centering
\includegraphics[width=0.4\linewidth]{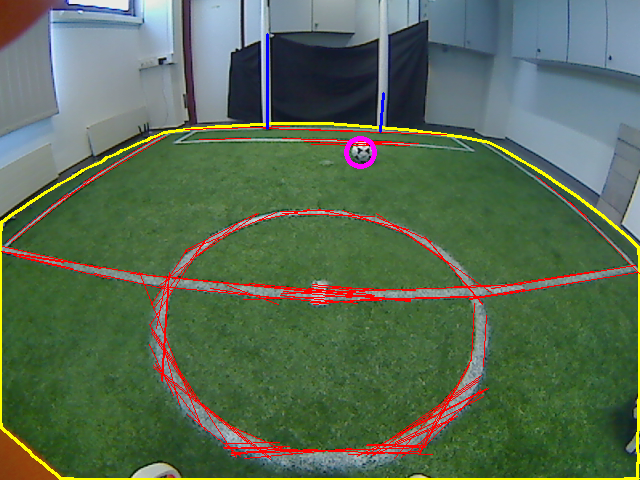}\hspace{0.019\linewidth}
\includegraphics[width=0.4\linewidth]{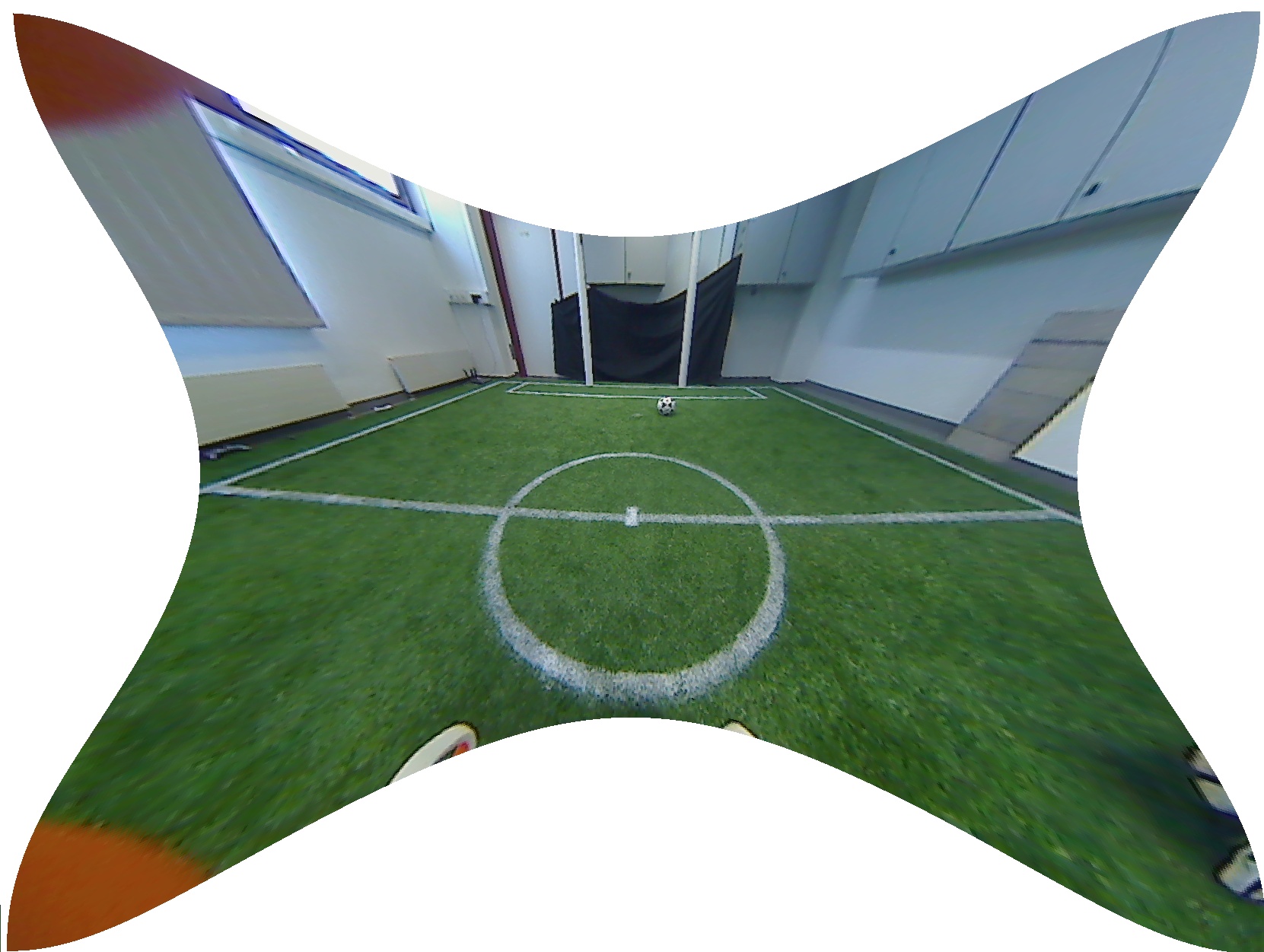}\\[2mm]
\includegraphics[width=0.4\linewidth,height=40mm]{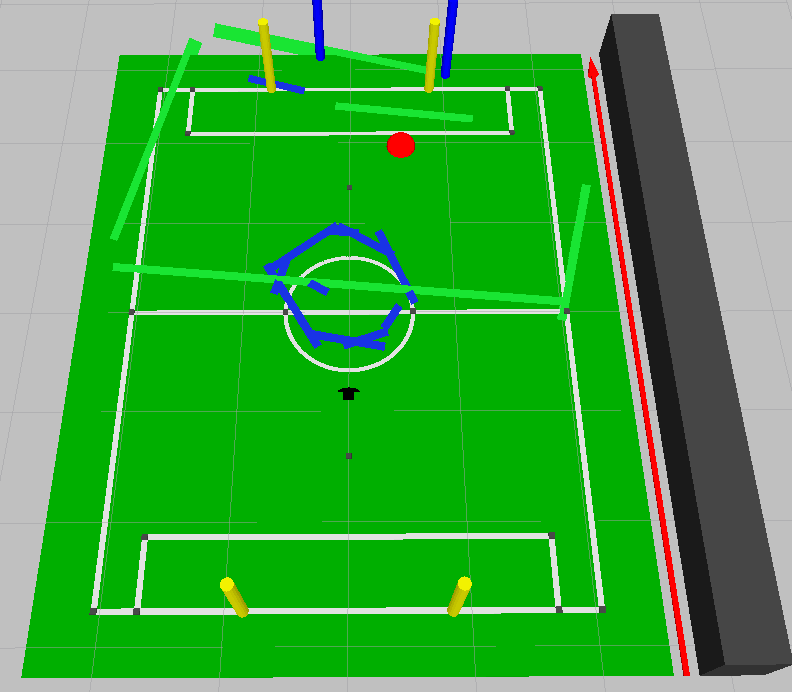}\hspace{0.02\linewidth}
\includegraphics[width=0.4\linewidth,height=40mm]{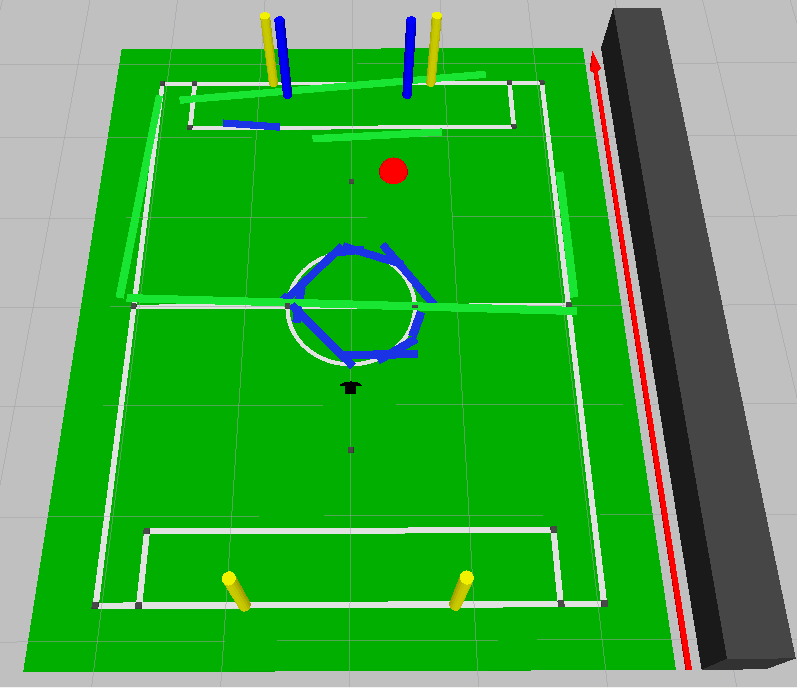}}
\caption{Top: A captured image (left) with ball (pink circle), field line (red 
lines), field boundary (yellow lines), and goal post (blue lines) detections 
annotated, and the corresponding raw captured image with undistortion applied 
(right). Bottom: Projected ball, field line and goal post detections before 
(left) and after (right) kinematic calibration.}
\figlabel{vision_output}
\vspace{-3ex}
\end{figure}
\ssubsection{State Estimation}

State estimation is a vital part of virtually any system that utilises 
closed-loop control. The 9-axis IMU on the microcontroller board is used to 
obtain the 3D orientation of the robot relative to its environment through the 
means of a nonlinear passive complementary filter \cite{Allgeuer2014}. This 
filter returns the full 3D estimated orientation of the robot with use of a 
novel way of representing orientations---the \term{fused angles} representation 
\cite{Allgeuer2015}. An immediate application of the results of the state 
estimation is the fall protection module, which disables the torque in order to 
minimise stress in all of the servos if a fall is imminent.

\ssubsection{Actuator Control}

As with most robots, motions performed by the \iguhop are dependent on the 
actuator's ability to track their set position. This is influenced by many 
factors, including battery voltage, joint friction, inertia and load. To 
minimise the effects of these factors, we apply feed-forward control to the 
commanded servo positions. This allows the joints to move in a compliant way, 
reduces servo overheating and wear, increases battery life, and reduces the 
problems posed by impacts and disturbances \cite{Schwarz2013a}. The vector of 
desired feed-forward output torques is computed from the vectors of commanded 
joint positions, velocities and accelerations using the full-body inverse 
dynamics of the robot, with help of the Rigid Body Dynamics Library. Each servo 
in the robot is configured to use exclusively proportional control. Time-varying 
dimensionless effort values on the unit interval $[0,1]$ are used per joint to 
interpolate the current proportional gain.

\ssubsection{Gait Generation}

The gait is formulated in three different pose spaces: joint space, abstract 
space, and inverse space. The \term{joint space} simply specifies all 
joint angles, while the \term{inverse space} specifies the Cartesian coordinates 
and quaternion orientations of each of the limb end effectors relative to the 
trunk link frame. The \term{abstract space} however, is a representation that 
was specifically developed for humanoid robots in the context of walking and 
balancing.

The walking gait is based on an open loop central pattern generated core that is 
calculated from a gait phase angle that increments at a rate proportional to the 
desired gait frequency. This open loop gait extends the gait of our previous 
work \cite{Missura2013a}. Since then, a number of simultaneously operating basic 
feedback mechanisms have been built around the open loop gait core to stabilise 
the walking \cite{Allgeuer2016a}. The feedback in each of these mechanisms 
derives from the fused pitch and fused roll state estimates, and adds corrective 
action components to the central pattern generated waveforms in both the 
abstract and inverse spaces \cite{Allgeuer2015}.

\ssubsection{Motions}

\begin{figure}[!t]
\parbox{\linewidth}{\centering\includegraphics[width=0.75\linewidth]{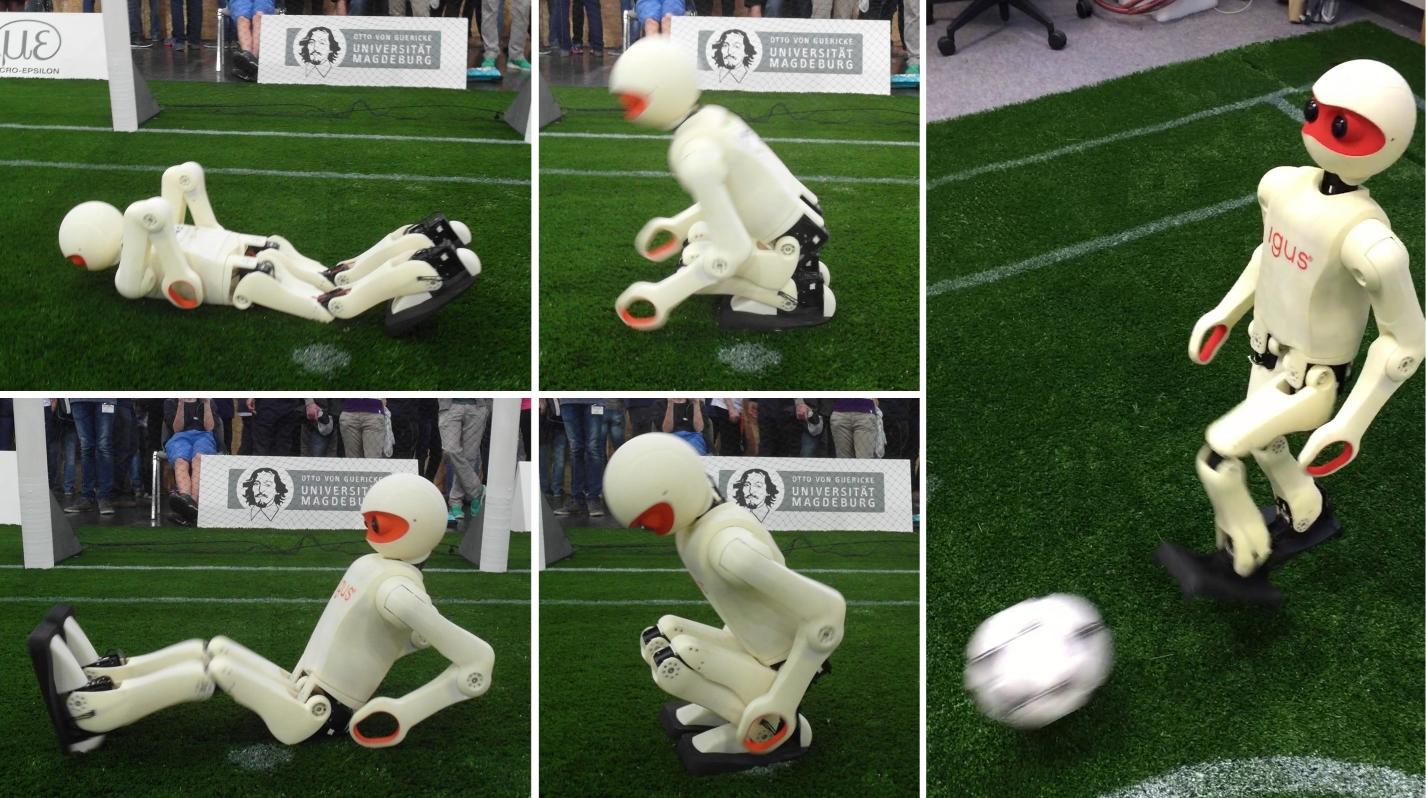}}
\caption{Dynamic get-up motions, from the prone (top row) and supine (bottom 
row) lying positions, and a still image of the dynamic kick motion.}
\figlabel{P1_getup}
\vspace{-3.5ex}
\end{figure}

Often there is a need for a robot to play a particular pre-designed motion. This 
is the task of the motion player, which implements a nonlinear keyframe 
interpolator that connects robot poses and smoothly interpolates joint positions 
and velocities, in addition to modulating the joint efforts and support 
coefficients. This allows the actuator control scheme to be used meaningfully 
during motions with changing support conditions. To create and edit the motions, 
a trajectory editor was developed for the \iguhop. All motions can be edited in 
a user-friendly environment with a 3D preview of the robot poses. We have 
designed numerous motions including kicking, waving, balancing, get-up, and 
other motions. A still image of the kicking motion is shown in \figref{P1_getup} 
along with the get-up motions of the \iguhopp, from the prone and supine 
positions.

\ssection{Reception}

To date we have built seven \iguhop{}s, and have demonstrated them at the 
RoboCup and various industrial trade fairs. Amongst others, this includes 
demonstrations at Hannover Messe in Germany, and at the International Robot 
Exhibition in Tokyo, where the robots had the opportunity to show their 
interactive side (see \figref{iguhop_interaction}). Demonstrations ranged from 
expressive and engaging looking, waving and idling motions, to visitor face 
tracking and hand shaking. The robots have been observed to spark interest and 
produce emotional responses in the audience.

Despite the recent design and creation of the platform, work groups have already 
taken inspiration from it, or even directly used the open-source hardware or 
software. A good example of this is the Humanoids Engineering \& Intelligent 
Robotics team at Marquette University with their MU-L8 robot 
\cite{stroud2013mu}. In their design they combined both an aluminium frame from 
the \nop and 3D printed parts similar to those of the \iguhop, as well as using 
ROS-based control software inspired by our own. A Japanese robotics business 
owner, Tomio Sugiura, started printing parts of the \iguhop on an FDM-type 3D 
printer with great success. Naturally, the platform also inspired other humanoid 
soccer teams, such as the WF Wolves \cite{taschwf}, 
to improve upon their own robots. The \nop, which was a prototype for the 
\iguhop, has been successfully used in research for human-robot interaction 
research at the University of Hamburg \cite{barros2014real}. We recently sold a 
set of printed parts to the University of Newcastle in Australia and await 
results of their work.

In 2015, the robot was awarded the first RoboCup Design Award, based on criteria 
such as performance, simplicity and ease of use. At RoboCup 2016, the platform 
also won the first International HARTING Open Source Prize, and was a 
fundamental part of the winning TeenSize soccer team. These achievements confirm 
that the robot is welcomed and appreciated by the community.

\begin{figure}[!t]
\parbox{\linewidth}{\centering
\includegraphics[height=4.1cm]{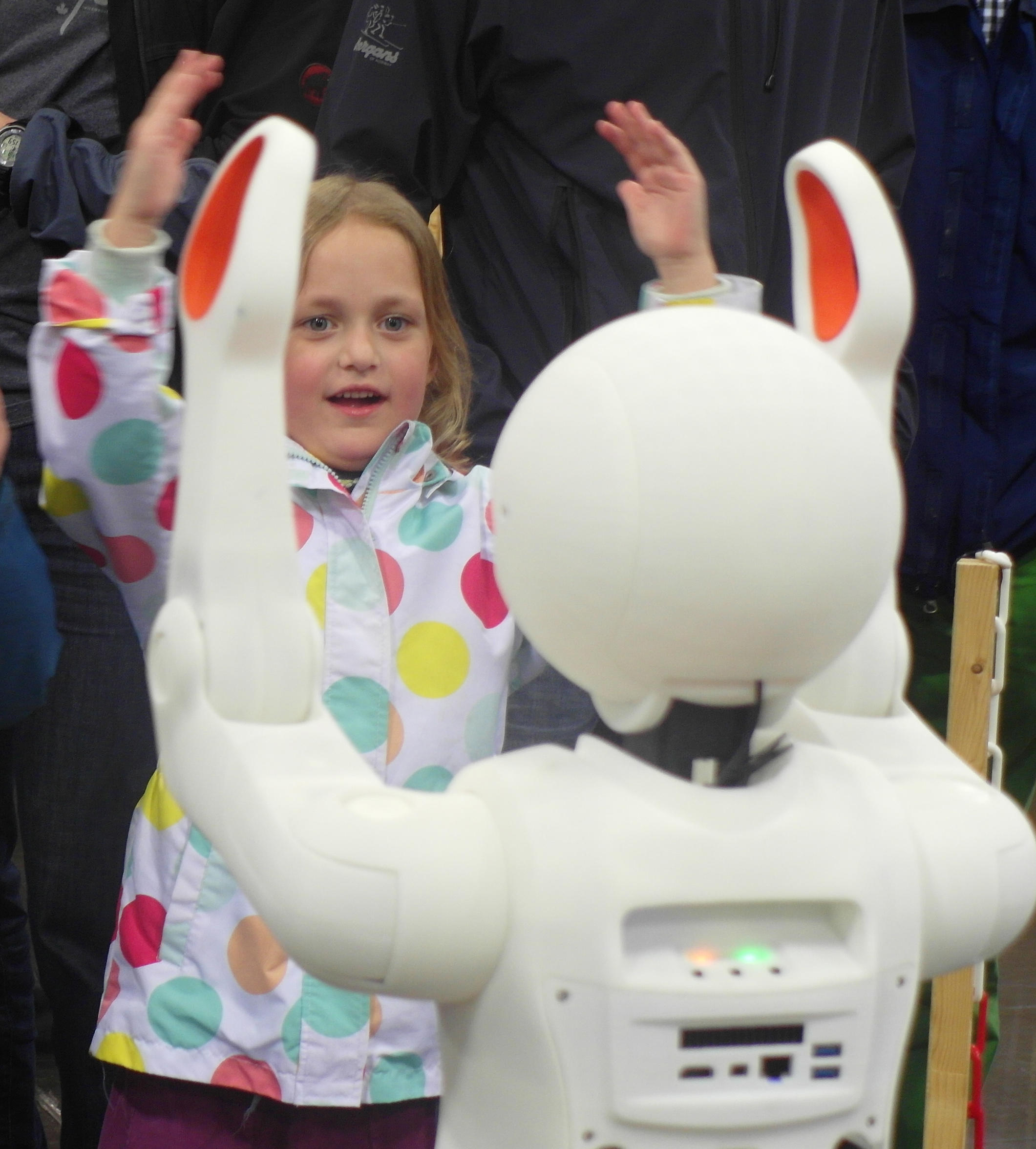}\hspace{0.02\linewidth}
\includegraphics[height=4.1cm]{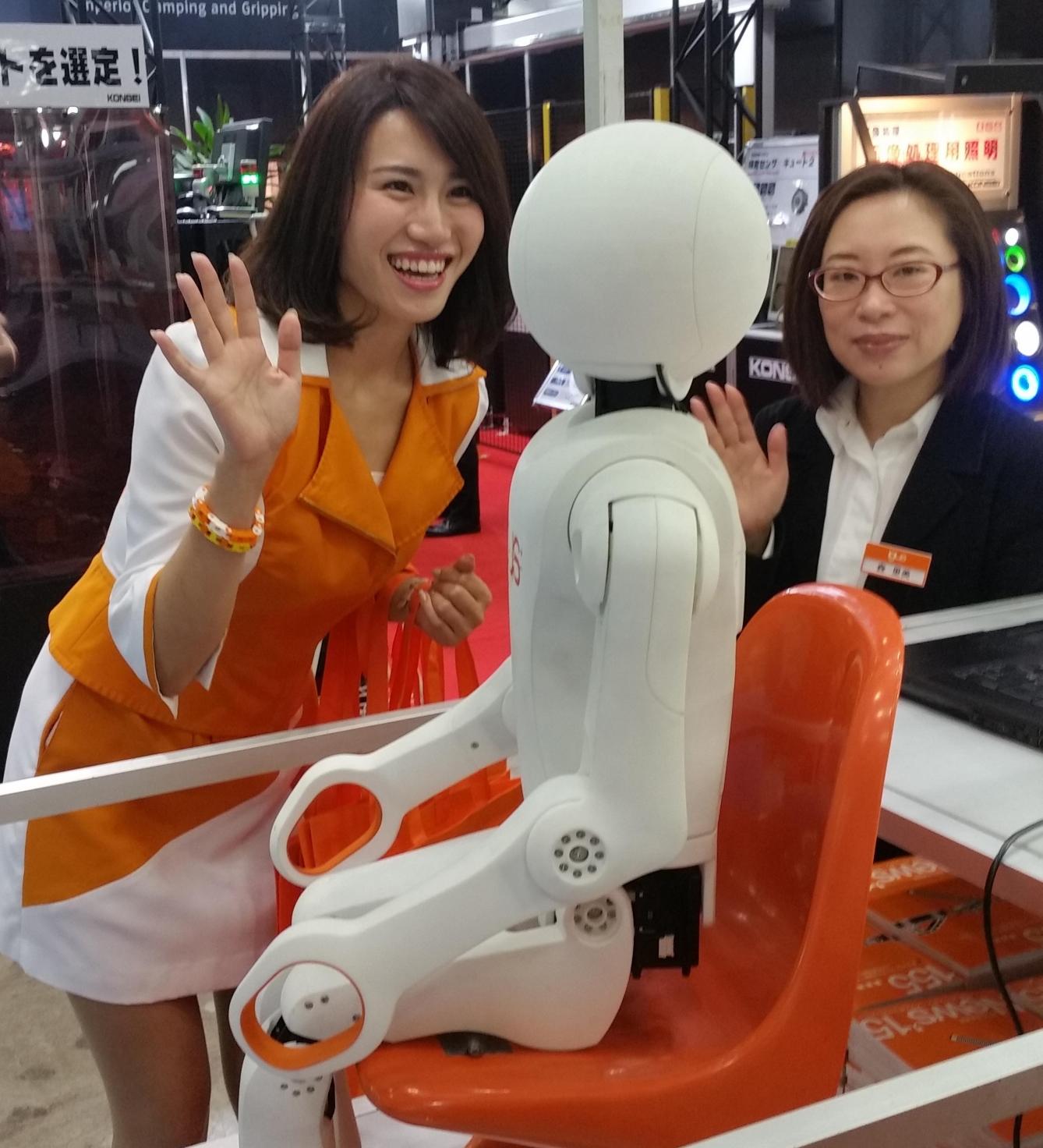}\hspace{0.02\linewidth}
\includegraphics[height=4.1cm]{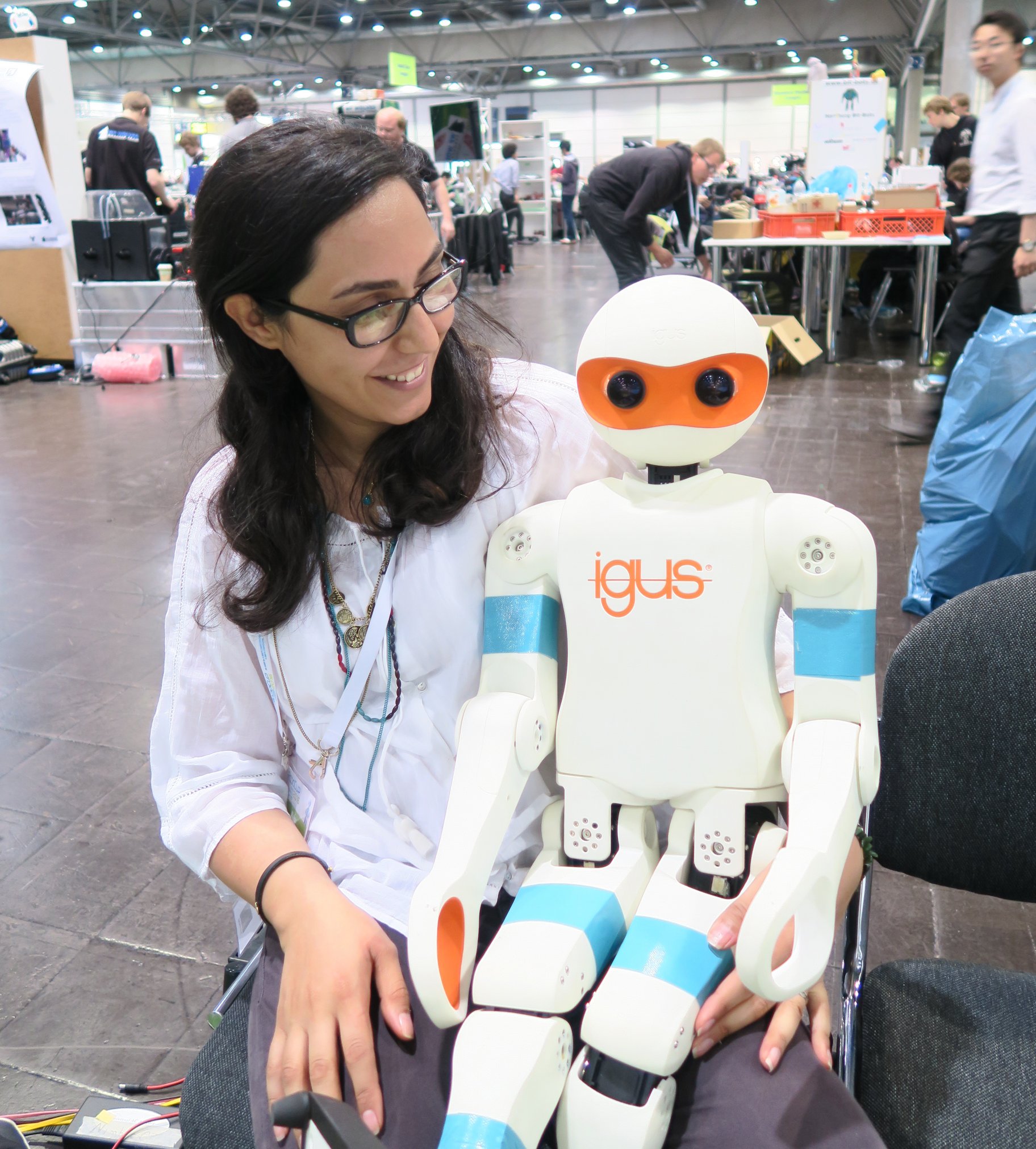}}
\caption{Example human interactions with the \iguhop, including waving to 
children (left), and face tracking (middle).}
\figlabel{iguhop_interaction}
\vspace{-3ex}
\end{figure}
\ssection{Conclusions}

Together with \igus GmbH, we have worked for three years to create and improve 
upon an open platform that is affordable, versatile and easy to use. The \iguhop 
provides users with a rich set of features, while still leaving room for 
modifications and customisation. We have released the hardware in the form of 
print-ready 3D CAD files\footnote{Hardware: 
\url{https://github.com/igusGmbH/HumanoidOpenPlatform}}, and uploaded the 
software to GitHub\footnote{Software: 
\url{https://github.com/AIS-Bonn/humanoid_op_ros}}. We hope that it will benefit 
other research groups, and encourage them to publish their results as 
contributions to the open-source community.

\ssections{Acknowledgements}
\footnotesize
We acknowledge the contributions of \igus GmbH to the project, in particular the 
management of Martin Raak towards the robot design and manufacture. This work 
was partially funded by grant BE 2556/10 of the German Research Foundation 
(DFG).

\bibliographystyle{ieeetr}
\bibliography{ms}

\end{document}